\documentclass{article}
\usepackage{arxiv}
\usepackage[utf8]{inputenc} 
\usepackage[T1]{fontenc}    
\usepackage{hyperref}       
\usepackage{url}            
\usepackage{booktabs}       
\usepackage{amsfonts}       
\usepackage{nicefrac}       
\usepackage{microtype}      
\usepackage{lipsum}		
\usepackage{graphicx}
\usepackage{doi}
\newcommand{\minitab}[2][l]{\begin{tabular}{#1}#2\end{tabular}}
\def\ie{\emph{i.e.}}

\usepackage{csquotes}
\usepackage{multirow}

\title{Low-Energy Convolutional Neural Networks (CNNs) using Hadamard Method}

\author{ \href{https://orcid.org/0000-0002-1866-6092}{\includegraphics[scale=0.06]{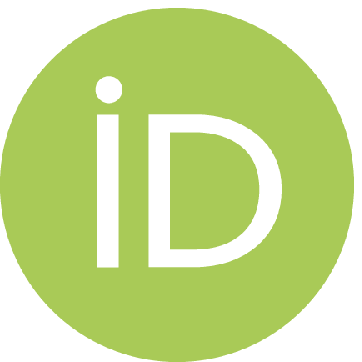}\hspace{1mm}Varun Mannam}\thanks{Varun Mannam is with the Department of Electrical Engineering, University of Notre Dame, Notre Dame, IN, 46556 USA.} \\
	Department of Electrical Engineering\\
	University of Notre Dame\\
	Notre Dame, IN 46556 \\
	\texttt{vmannam@nd.edu} \\
}

\renewcommand{\headeright}{A preprint}
\renewcommand{\shorttitle}{Low-energy CNNs using Hadamard method}

\begin{document}
\maketitle
\thispagestyle{plain}
\begin{abstract}
The growing demand for the internet of things (IoT) makes it necessary to implement computer vision tasks such as object recognition in low-power devices. Convolutional neural networks (CNNs) are a potential approach for object recognition and detection. However, the convolutional layer in CNN consumes significant energy compared to the fully connected layers. To mitigate this problem, a new approach based on the Hadamard transformation as an alternative to the convolution operation is demonstrated using two fundamental datasets, MNIST and CIFAR10. The mathematical expression of the Hadamard method shows the clear potential to save energy consumption compared to convolutional layers, which are helpful with BigData applications. In addition, to the test accuracy of the MNIST dataset, the Hadamard method performs similarly to the convolution method. In contrast, with the CIFAR10 dataset, test data accuracy is dropped (due to complex data and multiple channels) compared to the convolution method. Finally, the demonstrated method is helpful for other computer vision tasks when the kernel size is smaller than the input image size.
\end{abstract}



\section{INTRODUCTION}
Recently there has been a massive demand for low-energy internet of things (IoT) devices. Such devices involve applications like computer vision and image analysis that often require artificial neural networks (ANNs). Since energy is a constraint in these devices, there is a need to implement ANNs using energy-efficient techniques. This paper demonstrates the design and implementation of the Hadamard method in CNNs to address this challenge.

In ANN, deep learning uses a cascade of multiple layers for feature extraction \cite{Goodfellow:2016:DL:3086952}. Convolutional neural networks (CNNs) are specific deep learning architecture that enhances state-of-the-art performance for various image tasks, such as classification, object detection, and pattern recognition \cite{deveci2018energy}. A CNN consists of an input layer, an output layer, and multiple hidden layers. The hidden layers of a CNN typically consist of convolutional layers, pooling layers, and fully connected layers. Convolutional layers apply a convolution operation to the input, followed by a non-linear activation function. 

In CNNs, convolution takes the majority of the energy. Specifically, convolution with multiple layers consumes more than 90\% of the total energy resources \cite{abtahi2018accelerating}. In \cite{cnnoptimfft}, the energy consumption in 16-bit floating-point (FP) addition is 0.45 pJ, whereas 16-bit FP multiplication is 1.1 pJ. Similarly, for the 32-bit FP, energy consumption for multiplication and addition are 1.0 pJ and 4.5 pJ, respectively. Table \ref{table:energy} shows the number of multiplications and additions used in convolution for an image of size $N \times N$ and kernel of size $F \times F$. These computations are enormous when the kernel size is large, proving that the present multi-layer CNNs with convolution are not feasible with low-power devices. This requirement motivates us to find an elegant solution for implementing CNNs in energy-constrained devices. 
\subsection{Motivation}
In CNNs, we can use pruning and quantization of weights to reduce energy resources and also reduce accuracy/precision. When the kernel size is large, we can save more energy if the convolution is performed in the transform domain, where convolution is replaced with element-wise multiplication \cite{cnnoptimfft}. However, converting back and forth to the transform domain is expensive in this method, since it involves matrix multiplications. We propose using the Hadamard method to reduce the computation by replacing convolution with element-wise multiplication. In addition, the Hadamard method uses addition and subtraction instead of matrix multiplications in converting to the transform domain. Hadamard transform has been implemented on various tasks, such as data encryption and compression, to extract principal components and feature extraction. 

\textbf{Hadamard method:} In each convolutional layer, input from the previous stage and kernel are passed through the Hadamard transform \cite{orthotransform}, element-wise multiplied, and then passed through the inverse Hadamard transform, to bring it back to the spatial domain. In our demonstrated approach, we use Walsh-Hadamard transform (WHT), and inverse Walsh-Hadamard transform (IWHT) for faster implementation \cite{moiz1991fast}.
\subsection{Related work}
Different methods are proposed in the literature to reduce the energy resources in the convolutional layer. Binary weights network (BWN) \cite{xnornet} can compute the convolution with simple addition and subtraction. However, this method lacks accuracy. In Hadamard input Network (HIN) \cite{deveci2018energy}, the network is modified to use compressed images with the Hadamard transform instead of utilizing the original images as network inputs. The main contribution in reference \cite{deveci2018energy} is to use a combination of BWN and HIN called Binary weight network with Hadamard input (BWNHI) which uses less energy. Our approach extends this idea by using the Hadamard method in each convolutional layer.

The convolution operation can be performed in an orthogonal transform domain. For example, fast Fourier transform (FFT) can be used to reduce resources compared to convolution \cite{orthotransform}. In the FFT method \cite{cnnoptimfft}, in every convolutional layer, input and kernel are transformed to the frequency domain, element-wise multiplied, and finally converted back to the spatial domain using the inverse fast Fourier transform (IFFT). FFT domain multiplication results in the same values as performing convolution in the spatial domain. The non-linearity activation is applied in the spatial domain after convolution operation in each convolutional layer. In the FFT method, the convolution output is complex, whereas it is real in the Hadamard method. This reduces energy consumption, as shown in Table \ref{table:energy}. Additionally, in the Hadamard method, multiplications during the transformation of input and kernel using WHT are replaced by addition and subtraction, since the Hadamard coefficients are $\pm{1}$ \cite{moiz1991fast}. In all methods, the number of multiplications used in back-propagation is twice that of forwarding propagation. These results show that the Hadamard method is more energy efficient than convolution for large kernel sizes (please see Section.~\ref{energy_results}). \newline


\begin{table}[!ht]
\renewcommand{\arraystretch}{1.5}
\centering
\begin{tabular}{ccc}\hline
\centering
\textbf{Method} & \textbf{Multiplications} & \textbf{Additions} \\ \hline
Convolution & $N^2 F^2$ & $N^2 F^2-1$  \\
FFT \cite{mathieu2013fast} & $3N^2 \log{}N^2 + 4N^2$ & $3N^2 (N^2-1) +2N^2$  \\
\renewcommand{\arraystretch}{1.5}
\minitab[c]{\shortstack{\textbf{Hadamard method} \\\textbf{(Ours)}}} &  $N^2$ &  $3N^2 (N^2-1)$\\ \hline
\end{tabular} 
\caption{Number of computations to perform a simple convolution operation of input feature size of $N$ with a kernel of size $F$ using orthogonal domains \cite{cnnoptimfft}. Here $N$ indicates input feature size (assuming a single channel) and $F$ indicates kernel size. }\label{table:energy}
\end{table}

\subsection{Contributions}
In this report, the Hadamard method is used instead of convolution to extract the features in the images. The CNNs using the Hadamard method are implemented to show the effect of reduced energy consumption. In single-layer and three-layer CNNs, the performance of the Hadamard method is compared with convolution for MNIST \cite{lecun-mnisthandwrittendigit-2010} and CIFAR10 \cite{krizhevsky2009learning} datasets. In this paper, Section \ref{sec2} explains the Hadamard method with WHT. The CNNs architecture with the Hadamard method is given in Section \ref{sec3}. Section \ref{sec4} discusses the simulation results of CNNs with the Hadamard method and convolution on MNIST and CIFAR10 datasets including energy saving for single channel and multi-channel features with various kernel sizes. Conclusions and future work are given in Section \ref{sec5}. 

\section{Methodology} \label{sec2}
Hadamard coefficients are arranged either in ascending order or by using the Walsh Hadamard sequence \cite{moiz1991fast}. The first representation of the Hadamard kernel is convenient for representing the sequence order, whereas the second approach is simple for formalization. The Hadamard transform has the recursive property for any order (\ie, any order in which Hadamard transform coefficients are derived from its previous sequence of Hadamard transform coefficients) and is given by 
\begin{equation}
\centering
H(2N) = \left[ {\begin{array}{*{20}{c}}
{H(N)}&{H(N)} \\
{H(N)}&{ - H(N)}\end{array}} \right] \label{eq:11} 
\end{equation}
where $H(2)$ is given by
$ H(2) = \left[ {\begin{array}{*{20}{c}}
1&1\\
1&{ - 1} \end{array}} \right].$
The simplified version of WHT and IWHT of input image $X$ and Hadamard transform output $Y$ is given by
\begin{equation}
\centering
Y = HXH^T\label{eq:2} 
\end{equation}
\begin{equation}
\centering
X = \frac{1}{N^2}H Y {H^T}\label{eq:5} 
\end{equation}

where $H$ and $H^T$ are $N \times N$ Hadamard matrices and $N$ is the width and height of the image. The IWHT is similar to WHT except for the scaling factor. In the Hadamard method, we know image indices that are multiplied with kernel indices. To simplify further, using WHT, the convolution of image $x$ and kernel $h$ using the Hadamard method is given by 
\begin{equation}
x \star h [k,l] = \sum\limits_m {\sum\limits_p {x[m,p]h[k \oplus m,l \oplus p]} } \label{eq:3}
\end{equation}
where each index ($k$, $l$, $m$, $p$) range from 0 to $2^n-1$, $n$ number of bits required to represent each indices, $\star$ indicates the Hadamard method convolution and $\oplus$ represents the binary XOR operation \cite{mainreflink}. In the same way, the convolution of image $x$ and kernel $h$ is given by 
\begin{equation}
x * h [k,l] = \sum\limits_m {\sum\limits_p {x[m,p]h[k - m,l - p]} } \label{eq:4}
\end{equation}
where $m$, $p$ are the dimensions of image, $k$, $l$ are the dimensions of the convolution output and $*$ indicates convolution. Here the kernel dimensions are much smaller than the image dimension, and the multiplication will happen on the small image section. The sum is over all the tuples of $(m,p)$ that lead to valid subscripts for $x[m,p]$ and $h[k - m,l - p]$. Otherwise, those values are considered zeros in the summation. From Eq. {\ref{eq:3}} and Eq. {\ref{eq:4}}, the two results of the convolution and the Hadamard method are different. Still, their structure is similar in terms of the multiplication coefficients. This similarity suggests that we can extract the features of the images with the Hadamard method, similar to convolution in CNNs.   

\section{Network Architecture} \label{sec3}
\begin{figure}[h!]
\centering
\includegraphics[width=0.5\linewidth]{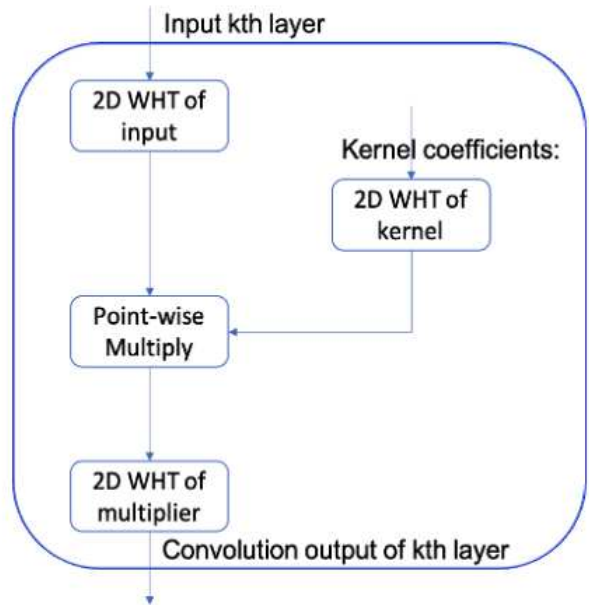}
\caption{Illustration of the Hadamard method to replace convolution operation in a convolutional layer using 2D WHT in the transform domain, where WHT coefficients are $\pm{1}$.} \label{fig:11}
\end{figure}

In supervised learning with CNN, kernel coefficients are learned using a training dataset to extract the features from images. In our experiment, we used multi-layer CNNs, which contain convolutional layers followed by a fully connected layer.  
\begin{figure}[h!]
\centering
\includegraphics[width=0.9\linewidth]{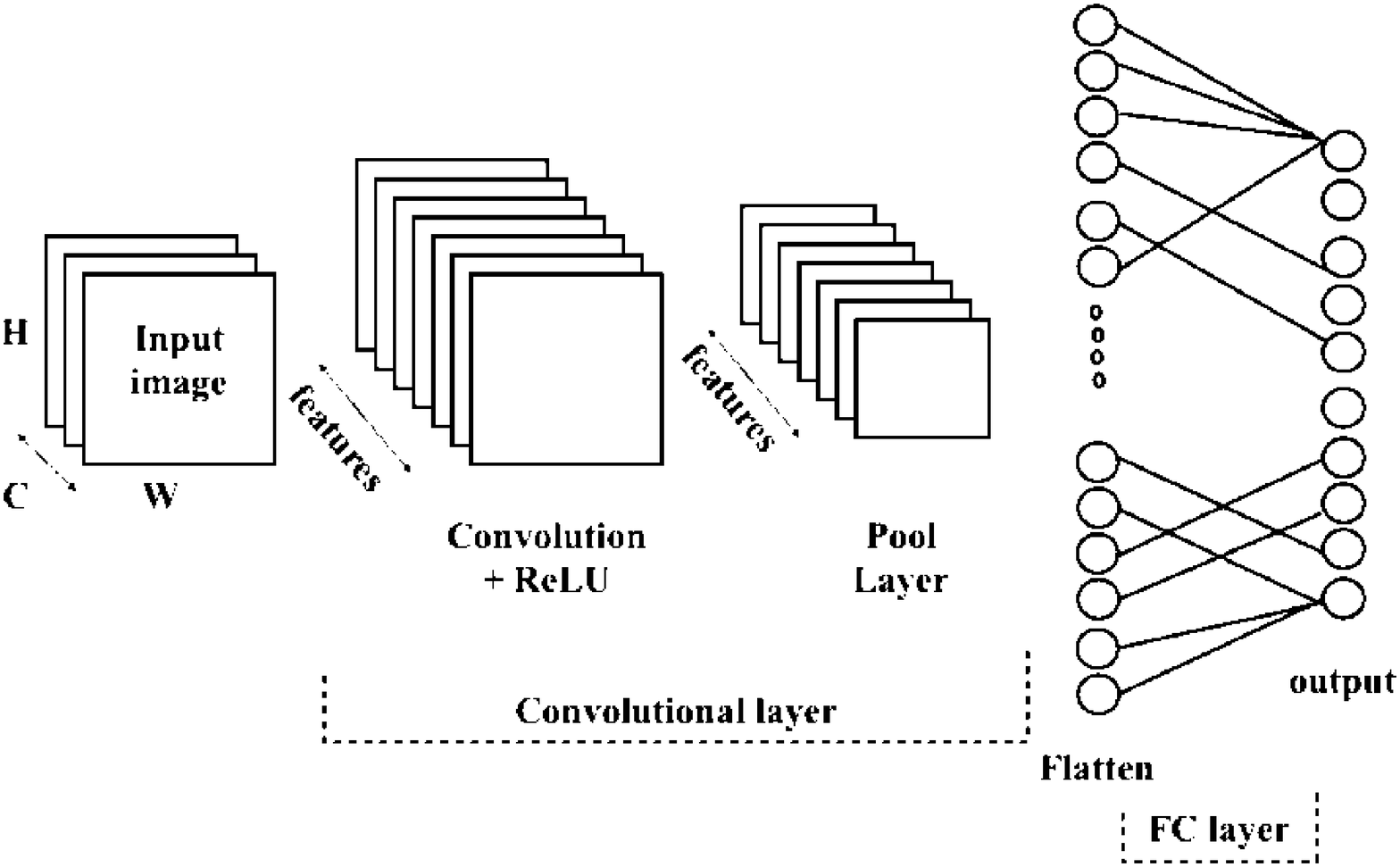}
\caption{Illustrating a simple CNN architecture which includes a convolutional layer (convolution operation/Hadamard method, non-linear activation), a pooling layer followed by a fully-connected layer.} \label{fig:1}
\end{figure} 
Fig.\ref{fig:1} shows a CNN with a single convolutional layer followed by a fully connected layer. Here convolutional layer contains convolution and non-linear activation, followed by the pooling layer. In our approach, we replaced the convolution with the Hadamard method in the convolution operation in the convolutional layer. In each convolutional layer, many kernels (number of features) extract different features from the input image. In the Hadamard method, an input image of size $C \times H \times W$ and a kernel are transformed using Eq. \ref{eq:2} and then element-wise multiplied. In Eq. \ref{eq:2}, we also check for the criteria that input image and kernel size are powers of 2; if not, zeros are appended at the end. The result is transformed to the spatial domain using Eq. \ref{eq:5}. The output of the Hadamard method is given to a rectified linear unit (ReLU) non-linear activation to extract all features. The output of non-linear activation is given to the pooling layer to extract the maximum/average value over the cluster of neurons. Lastly, the output from the pooling layer is flattened and then given to the fully connected layer to classify the image among the given classes. The output from the fully connected layer consists of the scores of each class for a given image, and the maximum scored class is considered the predicted class. The Hadamard method is extended to multi-layer CNNs to check the performance compared to convolution. A simple illustration of the Hadamard method in a convolutional layer is shown in Figure.~\ref{fig:11}.  

\section{Simulation Results}\label{sec4}
In this work, we consider four configurations: 
\begin{itemize}
    \item MNIST dataset with single-layer CNN
    \item CIFAR10 dataset with single-layer CNN
    \item MNIST dataset with three-layer CNN
    \item CIFAR10 dataset with three-layer CNN
\end{itemize}
For three-layer CNN architecture, three convolutional layers are cascaded. Simulation results show the performance of the Hadamard method for feature extraction with these datasets. We compare the test accuracy of the Hadamard method and convolution. The hyper-parameters are tuned to maximize the training accuracy (\ie, minimize the error between the predicted class and the labeled class). In our approach, the following parameters are tuned: batch size (BS), initial learning rate (ILR), weight decay (WD), and the number of output features of each convolutional layer. Additionally, we use \enquote{Adam} (adaptive moment estimation) optimizer and \enquote{ReLU} activation function. Since our task is image classification, we use cross-entropy as the loss function. We use adaptive learning rate (reduce learning rate by $10$ when the training loss is not decreased by a certain threshold over the last few epochs). At each convolutional layer, the input is batch-normalized before passing through the convolution block. In our network, the model parameters (like kernel coefficients and fully connected layer weights and bias) are initialized from a random sample set of a normal distribution, with zero mean and unit variance. Finally, we provide the results with the kernel sizes of $3 \times 3$, $5 \times 5$, and $7 \times 7$. For the energy calculations, we showed our results with larger kernel sizes. In our simulations, we tune the hyper-parameters for the kernel size $3 \times 3$, and we use these hyper-parameters for all the kernels in each configuration. The tuned hyper-parameters for all configurations are given in Table \ref{table:params} for the MNIST and CIFAR-10 datasets. \newline

\begin{table}[!ht]
\centering
\begin{tabular}{cccccc} \hline
\textbf{Dataset} & \textbf{CNN layers }& \textbf{H-ILR   } & \textbf{H-WD }    & \textbf{C-ILR }   &\textbf{ C-WD}     \\ \hline
MNIST   & 1          & 1.00E-04 & 1.00E-04 & 1.00E-04 & 1.00E-04 \\
        & 3          & 1.00E-04 & 1.00E-04 & 1.00E-04 & 1.00E-04 \\ \hline
CIFAR10 & 1          & 1.00E-03 & 1.00E-04 & 1.00E-03 & 1.00E-04 \\
        & 3          & 2.00E-03 & 1.00E-04 & 1.00E-03 & 1.00E-04 \\ \hline
\end{tabular}
\caption{Hyper-parameters for all configurations, where CNN networks with 1: single-layer CNN, 3: three-layers CNN, H-ILR: Hadamard method ILR, H-WD: Hadamard method WD, C-ILR: convolution ILR and C-WD: convolution WD. }\label{table:params}
\end{table}


\subsection{MNIST}\label{mnist}
MNIST is a dataset of hand-written digits from 0 to 9 grayscale (with a single channel as input) images of size $28 \times 28$ ($W \times H$) with 60000 training dataset and 10000 test dataset \cite{lecun-mnisthandwrittendigit-2010}.
\subsubsection{Single-layer CNN}\label{m1}
To match the number of parameters in each method, we disabled the bias in the convolution. The single-layer CNN parameters are kernel coefficients of the convolutional layer and weights and biases of the fully connected layer.

\begin{table}[!ht]
\centering
\begin{tabular}{cccc} \hline
\textbf{CNN layers} & \textbf{Kernel Size} & \textbf{Hadamard method (Ours)} & \textbf{Convolution method}\\ \hline
1          & 3×3         & 97.5                   & 98.34              \\
           & 5x5         & 97.95                  & 98.67              \\
           & 7x7         & 97.97                  & 98.83              \\ \hline
3          & 3x3         & 97.91                  & 99.27              \\ 
           & 5x5         & 98.28                  & 99.37              \\
           & 7x7         & 98.43                  & 99.38          \\ \hline   
\end{tabular}
\caption{MNIST test dataset accuracy (\%) for single and three-layer CNNs using convolution and Hadamard methods. CNN layers 1: single-layer CNN and 3: three-layers CNN.}\label{table:results1}
\end{table}


The results show that the test accuracy improved with an increase in kernel size due to more parameters being used to extract the features. The test accuracy in single-layer CNN for all kernels is given in Table \ref{table:results1}.

\subsubsection{Three-layer CNNs}\label{m3}
\begin{figure}[h!]
\centering
\includegraphics[width=0.6\linewidth]{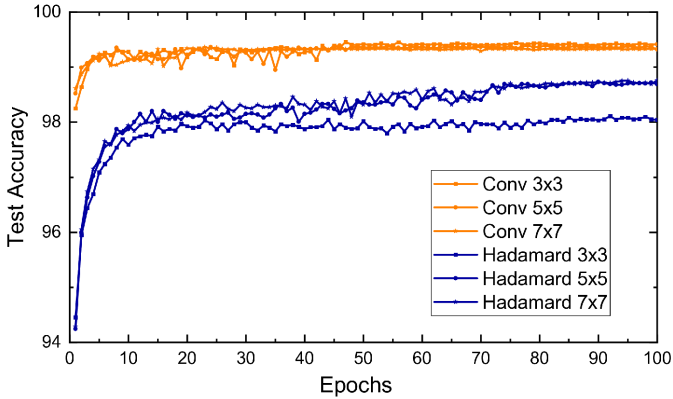}
\caption{Test accuracy of the MNIST dataset vs. epochs for three-layer CNN using the Hadamard and convolution methods.} \label{fig:3}
\end{figure} 
To improve network accuracy, we introduce more convolutional layers, which increase the number of parameters and extract more complex features at each layer compared to the previous layer. The test accuracy in three-layer CNN for all kernels is given in Table \ref{table:results1}, and we observe that when the number of convolutional layers increases, the test accuracy difference between Hadamard transform and convolution also increases. In this way, we show using the Hadamard method, with fewer computations, we got test-dataset accuracy similar to the traditional convolutional layers.

\subsection{CIFAR10}\label{cifar}
CIFAR10 dataset contains ten different classification images such as plane, cat, dog, ..etc with three input channels (C=3 represent R, G, B channels) of size $32 \times 32$ ($W \times H$). CIFAR10 dataset has 50000 training images, and 10000 test images \cite{krizhevsky2009learning}. The initial convolutional layer has kernels of size output features $\times$ input channels $\times$ kernel (\ie, out features $\times$ 3 $\times$ kernel). At the initial convolutional layer, each input channel is the convolution method (or convolution operation) with the same channel kernel, and all input channels are added together. This dataset uses a batch size (BS) of 20. When the learning rate is increased, the present model will overfit. Incorporating the batch-norm layers in the current architecture before the non-linear activation function improves the test accuracy in the CIFAR10 dataset (both in single-layer and three-layer) \cite{ioffe2015batch}. Adding batch norm after convolution method or Hadamard method leverages the training process to use a high learning rate that yields faster convergence. With a high learning rate, there is always an over-fitting issue and batch-norm works as a regularization technique. Adding dropout along with batch-norm at each layer provides better regularization. Dropout is another regularization method that randomly removes a few nodes/input channels with a certain probability over each batch size so that the model will not overfit. For CIFAR10, our network architecture is Hadamard method or convolution method, batch-normalization layers, and ReLU layer followed by a max-pool layer with dropout. 

\subsubsection{Single-layer CNN}\label{c1}
The tuned hyper-parameters are given in Table \ref{table:params}. The dropout value in our CIFAR10 dataset CNN model is set to 0.2 and 0.3 for Hadamard and convolution methods, respectively. The test accuracy using the CIFAR-10 dataset in single-layer CNN for all kernels is given in Table \ref{table:results2}. 

\begin{table}[!ht]
\centering
\begin{tabular}{cccc} \hline
\textbf{CNN layers} & \textbf{Kernel Size} & \textbf{Hadamard method (Ours)} & \textbf{Convolution method}\\ \hline
1          & 3×3         & 60.25                  & 64.06              \\
           & 5x5         & 61.43                  & 65.89              \\
           & 7x7         & 61.21                  & 66.08              \\ \hline
3          & 3x3         & 73.48                  & 82                 \\
           & 5x5         & 72.65                  & 81.97              \\
           & 7x7         & 71.08                  & 81.19             \\ \hline
\end{tabular}
\caption{CIFAR-10 test dataset accuracy (\%) for single and three-layer CNNs using convolution and Hadamard methods. CNN layers 1: single-layer CNN and 3: three-layers CNN.}\label{table:results2}
\end{table}


\subsubsection{Three-layer CNN}\label{c3}
\begin{figure}[h!]
\centering
\includegraphics[width=0.6\linewidth]{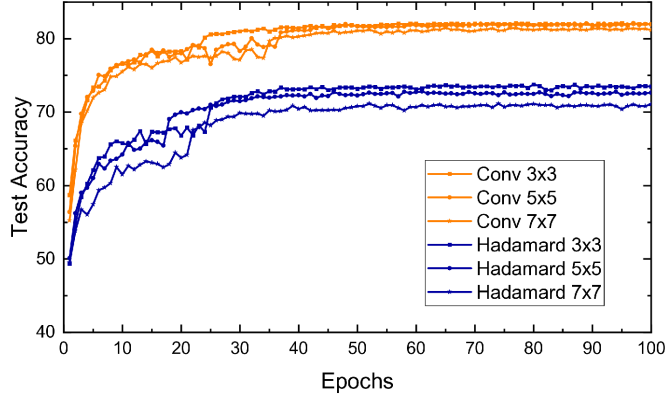}
\caption{Test accuracy of the CIFAR-10 dataset vs. epochs for three-layer CNN using the Hadamard and convolution methods.} \label{fig:5}
\end{figure} 
To improve the test accuracy, we increase the number of features by two after every convolutional layer with the CIFAR10 dataset. This is because, after each convolutional layer, the image size is reduced by half (because of the max-pool with a kernel size of $2 \times 2$). To extract more complex features at each successive convolutional layer, the number of output features doubles compared to the previous layer. The number of output features is 32, 64, and 128 in the first, second, and third convolutional layers, respectively. Further, the test accuracy can be improved by adding a second fully connected layer at the end. The ReLU activation in the first fully connected layer along with dropout is applied. In addition, in the Hadamard method and convolution method in all convolutional and fully connected layers, dropout is set to 0.2. 

\subsection{Energy results} \label{energy_results}
\begin{figure}[h!]
\centering
\includegraphics[width=0.7\linewidth]{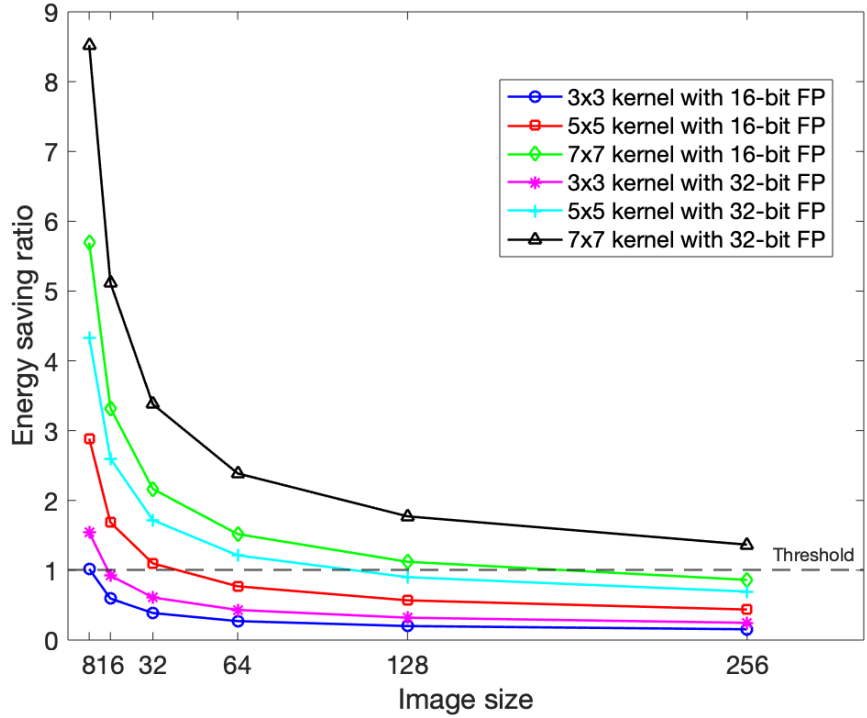}
\caption{Single convolutional layer computational energy saving ratio vs. single channel image size across kernels for 16-bit and 32-bit floating point multiplications.} \label{fig:6}
\end{figure} 
In this section, energy-saving calculations are shown for the Hadamard method when compared to the convolution method in a simple CNN structure. For example, for a single convolutional layer with a single channel image of size N$\times$N with a kernel size of F$\times$F, then the number of multiplications and additions required to generate convolution output is given in Table.~\ref{table:energy}. Consider, energy consumption for multiplication and addition are $E_m$ and $E_a$, then total energy consumption in convolution method is $N^2*F^2*E_m + N^2*(F^2-1)*E_a$. Similarly, for the Hadamard method, the energy consumption is $N^2*E_m + 3N^2*(N^2-1)*E_a$. The energy saving ratio is defined as the energy consumption of the convolution method to the energy consumption of the Hadamard method. The energy ratio is $\frac{F^2*\alpha+(F^2-1)}{\alpha+3(N^2-1)}$ where $\alpha$ is the ratio of $\frac{E_m}{E_a}$. For 16-bit and 32-bit FP, the $\alpha$ values are 2.44 and 4.5, respectively. To show the energy-saving ratio for a single channel image, we perform a simulation across different image and kernel sizes which is shown in Figure.~\ref{fig:6}. Here considered both 16-bit floating point (FP) and 32-bit floating point multiplication energy values. From the simulation, for the baseline configuration where the threshold is set to 1, the image is less than 128 for the typical kernels of 3$\times$3 or 5$\times$5 which is typically the same as most image patches in size.  

\begin{figure}[h!]
\centering
\includegraphics[width=0.7\linewidth]{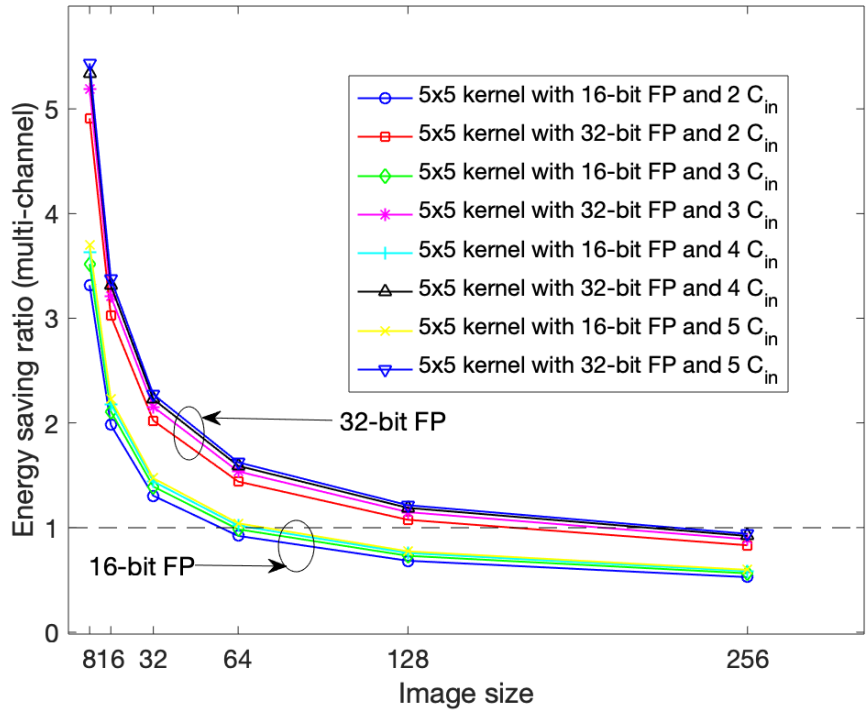}
\caption{Single convolutional layer computational energy saving ratio vs. multiple channels/features $C_{in}$ size across kernels for 16-bit and 32-bit floating point multiplications.} \label{fig:7}
\end{figure} 

However, most of the intermediate layers in CNNs are multi-channel features with the number of channels being $C_{in}$. The energy saving for multi-channel inputs is given by $\frac{F^2*(\alpha+1)C_{in}}{C_{in}*(\alpha+1+2(logN)^2)+(logN)^2)}$. The simulation result with various image sizes and kernel sizes is given in Figure.~\ref{fig:7}. In addition, various $C_{in}$ starting from 2 to 5 with floating point multiplications are shown with a baseline energy saving of 1. From Figure.~\ref{fig:7}, an image of size less than 128 provides energy saving using the Hadamard method across all kernels.

\subsection{Results discussion}
For kernel of $5 \times 5$, from the MNIST dataset, there is a 0.71\% gap in test accuracy between Hadamard and convolution methods in single-layer CNN, whereas the gap in accuracy increases to 1.09\% with three-layer CNN. MNIST is a simple dataset with grayscale images in which most of the features can be extracted using the Hadamard method. In contrast, the CIFAR-10 dataset is complex since 3-channel input images and for the CIFAR10 dataset, in single-layer CNN, there is a gap of 4.46\% in test accuracy between the Hadamard and convolution methods. This gap increases to 9.32\% for three-layer CNN between the methods. In the CIFAR10 dataset, the test accuracy gap is more due to image content and complex features. With our present architecture, the gap in test accuracy with three-layer CNNs between the Hadamard method and convolution methods can be reduced with proper hyperparameters. The main objective of the proposed method is to minimize computations to achieve the classification task on low-energy devices, such that the entire object classification task can be fit to low-power real-time devices such as mobiles. One exciting application is, proposed simple object detection networks perform inference using the Hadamard method where the multiplication operations are substantially reduced. 


\section{Conclusions and future work} \label{sec5}
\subsection{Conclusions}
In CNNs, convolution is energy expensive, and we identified an alternative method for feature extraction, namely the Hadamard method. We implemented the multi-layer CNNs with the Hadamard method and verified them with MNIST and CIFAR10 datasets. We observed that the MNIST dataset's Hadamard transform in CNN achieves performance similar to convolution (test accuracy difference is around 1\%). However, in the CIFAR10 dataset, the Hadamard method underperforms convolution (test accuracy difference is 9.32\%) since the dataset is complex. Initial results show that using the Hadamard method; we can extract the features in the images using few energy resources with reduced performance. 

\subsection{Future work}
The test accuracy can be improved in both datasets with operations (adjusting the positions of the coefficients) on the Hadamard method output such that the difference between Hadamard output and convolution is slight and by extending this idea to multi-layer CNNs by comparing the performance of the Hadamard method and convolution methods. Similar to Winograd CNNs \cite{winograd}, performing the Hadamard method without converting back to the spatial domain after each convolutional layer can lead to interesting observations.  

\bibliography{references} 

\begin{thebibliography}{10}

\bibitem{Goodfellow:2016:DL:3086952}
Ian Goodfellow, Yoshua Bengio, and Aaron Courville.
\newblock {\em Deep Learning}.
\newblock The MIT Press, 2016.

\bibitem{deveci2018energy}
T~Ceren Deveci, Serdar Cakir, and A~Enis Cetin.
\newblock Energy efficient hadamard neural networks.
\newblock {\em arXiv preprint arXiv:1805.05421}, 2018.

\bibitem{abtahi2018accelerating}
Tahmid Abtahi, Colin Shea, Amey Kulkarni, and Tinoosh Mohsenin.
\newblock Accelerating convolutional neural network with fft on embedded
  hardware.
\newblock {\em IEEE Transactions on Very Large Scale Integration (VLSI)
  Systems}, 26(9):1737--1749, 2018.

\bibitem{cnnoptimfft}
Artem Vasilyev.
\newblock Cnn optimizations for embedded systems and fft.
\newblock 2015.

\bibitem{orthotransform}
Anna U{\v{s}}{\'a}kov{\'a}, Jana Kotuliakov{\'a}, and Michal Zajac.
\newblock Using of discrete orthogonal transforms for convolution.
\newblock {\em Journal of Electrical Engineering}, 53(9-10):285--288, 2002.

\bibitem{moiz1991fast}
Saifuddin Moiz.
\newblock {\em Fast implementation of hadamard transform for object recognition
  and classification using parallel processor}.
\newblock PhD thesis, Ohio University, 1991.

\bibitem{xnornet}
Mohammad Rastegari, Vicente Ordonez, Joseph Redmon, and Ali Farhadi.
\newblock {XNOR-Net}: {I}magenet classification using binary convolutional
  neural networks.
\newblock {\em CoRR}, abs/1603.05279, 2016.

\bibitem{mathieu2013fast}
Michael Mathieu, Mikael Henaff, and Yann LeCun.
\newblock Fast training of convolutional networks through ffts.
\newblock {\em arXiv preprint arXiv:1312.5851}, 2013.

\bibitem{lecun-mnisthandwrittendigit-2010}
Yann LeCun and Corinna Cortes.
\newblock {MNIST} handwritten digit database.
\newblock 2010.

\bibitem{krizhevsky2009learning}
Alex Krizhevsky, Geoffrey Hinton, et~al.
\newblock Learning multiple layers of features from tiny images.
\newblock 2009.

\bibitem{mainreflink}
Math~Stack Exchange.
\newblock {Convolution-theorem-for-other-transforms}.
\newblock
  \url{https://math.stackexchange.com/questions/577491/convolution-theorem-for-other-transforms/}.

\bibitem{ioffe2015batch}
Sergey Ioffe and Christian Szegedy.
\newblock Batch normalization: Accelerating deep network training by reducing
  internal covariate shift.
\newblock In {\em International conference on machine learning}, pages
  448--456. PMLR, 2015.

\bibitem{winograd}
Xingyu Liu, Jeff Pool, Song Han, and William~J. Dally.
\newblock Efficient sparse-winograd convolutional neural networks.
\newblock {\em CoRR}, abs/1802.06367, 2018.

\end{thebibliography}
\bibliographystyle{unsrt} 

\end{document}